\documentclass[review]{elsarticle}

\usepackage{lineno}
\modulolinenumbers[5]
\usepackage{amsmath}
\usepackage{booktabs}
\usepackage{eqparbox}
\usepackage{amsmath,amsfonts,amssymb}
\usepackage{graphicx}
\usepackage[colorlinks=true, allcolors=blue]{hyperref}
\usepackage{float}
\usepackage[caption = false]{subfig}
\usepackage{siunitx}
\usepackage[table,xcdraw]{xcolor}
\usepackage{epstopdf}
\usepackage[colorinlistoftodos]{todonotes}
\usepackage{amsmath}

\journal{Journal of Visual Communication and Image Representation }

\begin{document}
\begin{frontmatter}

\title{ Convolutional Neural Networks: Ensemble Modeling, Fine-Tuning and Unsupervised Semantic Localization for Intraoperative CLE Images}
\author[a,b]{Mohammadhassan Izadyyazdanabadi\corref{correspondingauthor}}
\cortext[correspondingauthor]{Corresponding author:}
\ead{mizadyya@asu.edu}
\author[a,b,c]{Evgenii Belykh}
\author[b]{Michael Mooney} 
\author[b]{Nikolay Martirosyan}
\author[b]{Jennifer Eschbacher} 
\author[b]{Peter Nakaji}
\author[b]{Mark C. Preul}
\author[a]{Yezhou Yang}

\address[a]{Arizona State University, Tempe AZ 85281, USA}
\address[b]{Department of Neurosurgery, Barrow Neurological Institute, St Joseph's Hospital and Medical Center, Phoenix, AZ 85013}
\address[c]{Irkutsk State Medical University, Krassnogo vosstaniya 1, Irkutsk, Russia 664003}




\begin{abstract}
Confocal  laser  endomicroscopy  (CLE)  is an  advanced  optical  fluorescence technology undergoing assessment for applications in brain tumor surgery.  Many of the CLE images can be distorted and interpreted as nondiagnostic. However, just one neat CLE image might suffice for intraoperative diagnosis of the tumor. While manual  examination of thousands of nondiagnostic images during surgery would be impractical, this creates an opportunity for a model to select diagnostic images  for the  pathologists or  surgeons  review.   In  this  study,   we sought to develop  a deep learning  model to  automatically detect  the  diagnostic  images. We explored the effect of training regimes and ensemble modeling and localized histological  features  from diagnostic  CLE images.  The  developed  model could achieve higher agreement with the ground truth than the other human observers. With  the  speed  and  precision  of the  proposed  method,  it  has  potential to  be integrated into the operative  workflow in the brain  tumor  surgery.
\end{abstract}


\end{frontmatter}


\section{Introduction}
Handheld, portable Confocal Laser Endomicroscopy (CLE) is being explored in neurosurgery because of its ability to image histopathological features of tissue in real time \cite{belykh2016intraoperative,charalampaki2015confocal, foersch2012confocal,sanai2011intraoperative}. CLE provides cellular resolution imaging during brain tumor surgery and thus may provide the surgeon with precise histopathological information during tumor resection in order to interrogate regions that may harbor malignant or spreading tumor, especially at the tumor border. 

\begin{figure}[ht!]
\centering
\subfloat[]{\includegraphics[width = 1.7in]{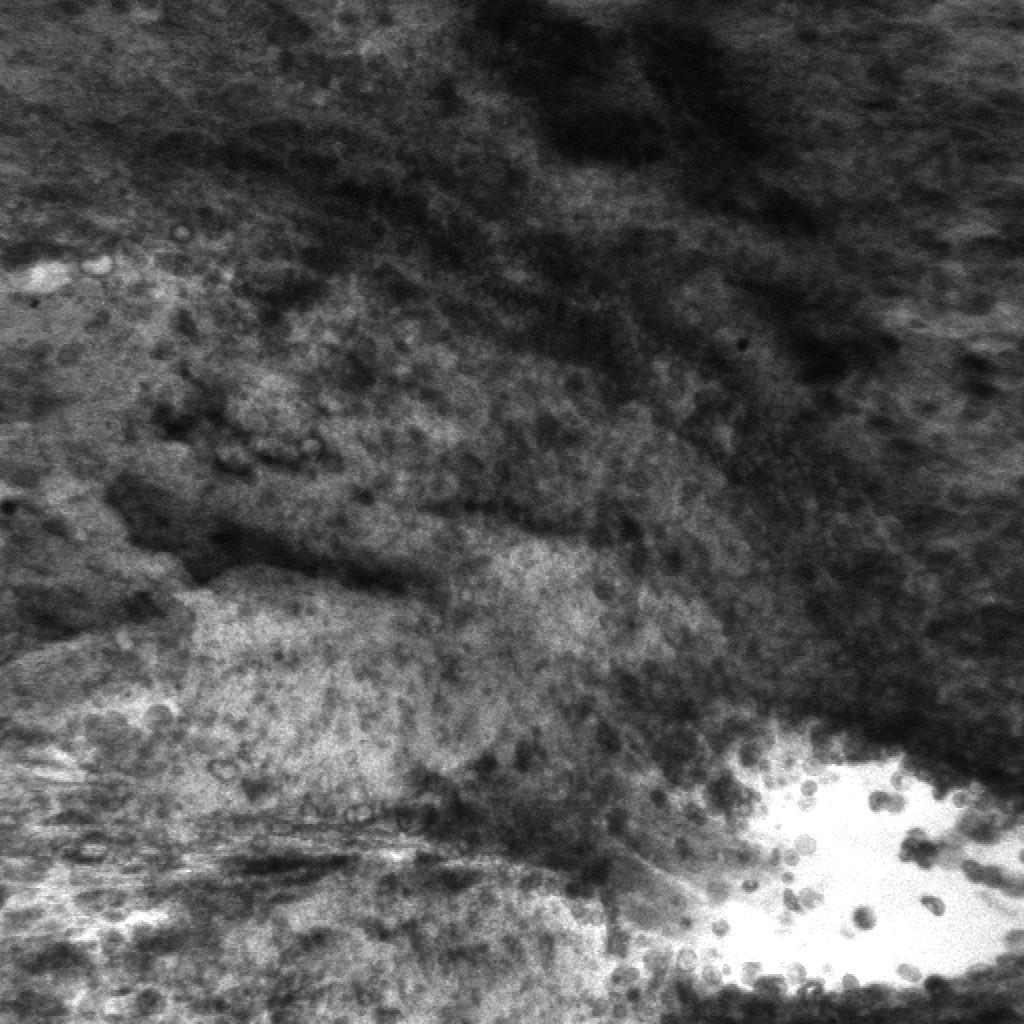}} ~
\subfloat[]{\includegraphics[width = 1.71in]{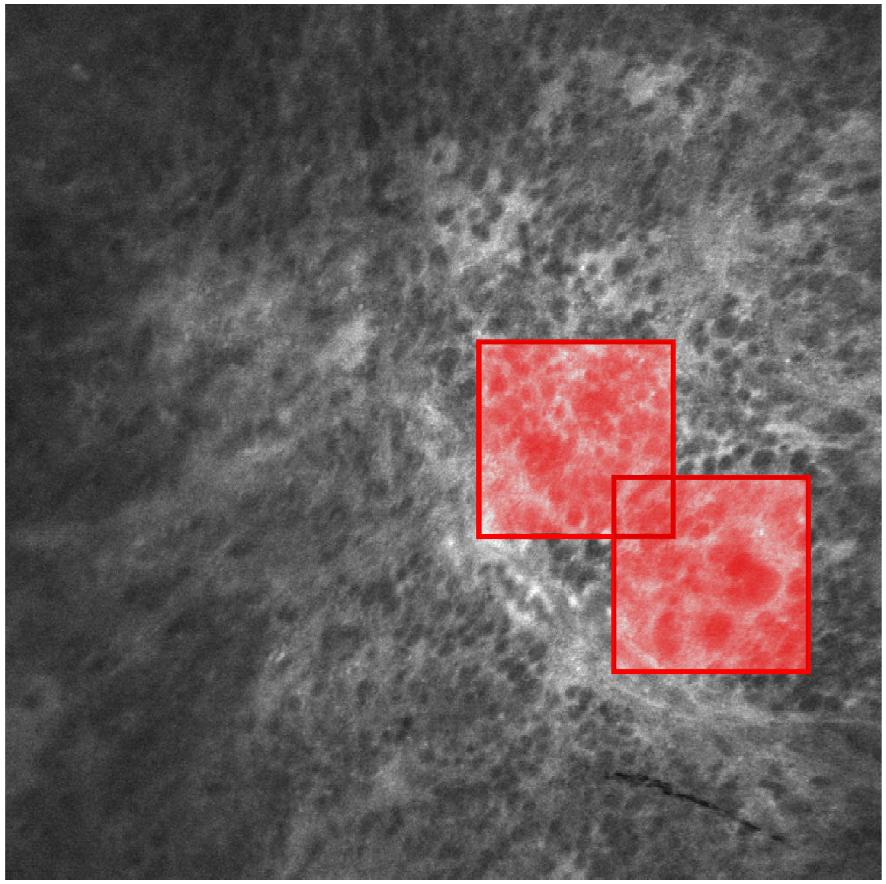}}\
\subfloat[]{\includegraphics[width = 1.7in]{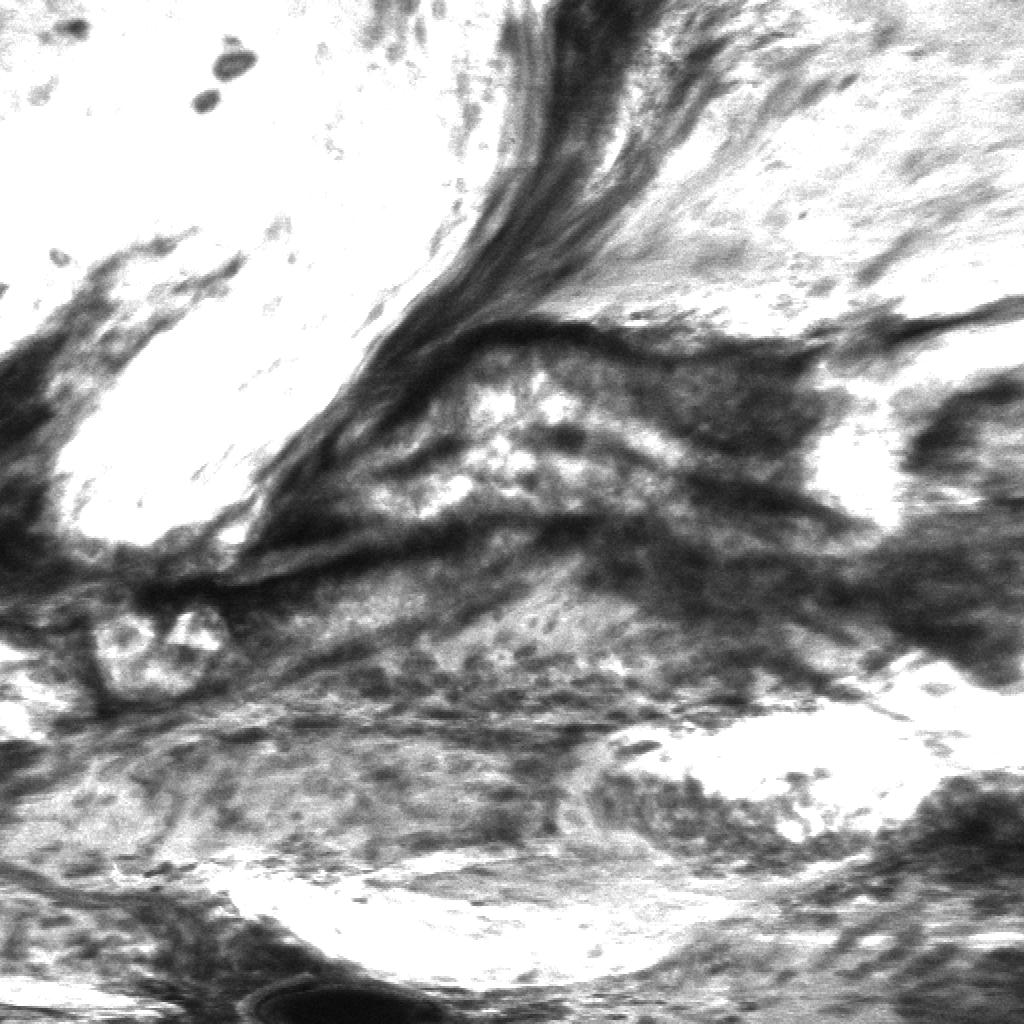}} ~
\subfloat[]{\includegraphics[width = 1.7in]{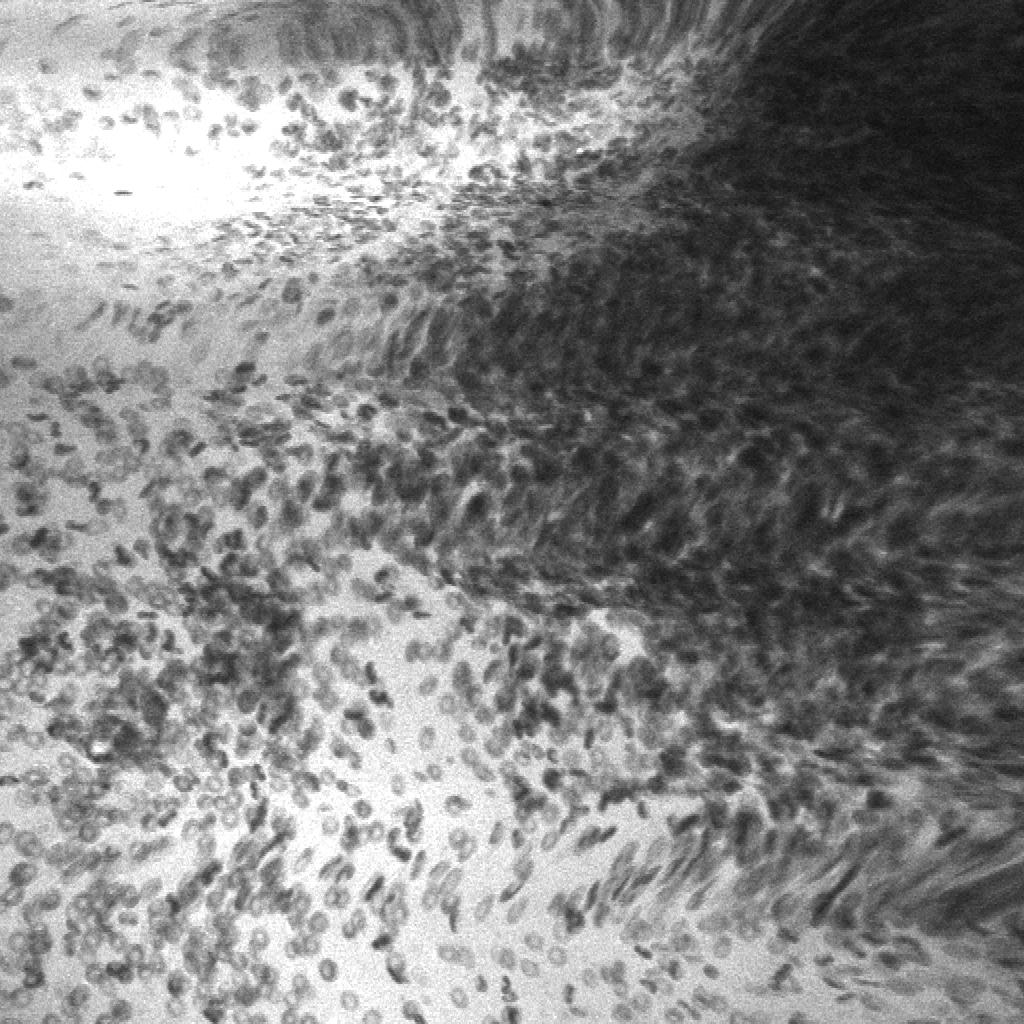}}
\caption{Diagnostic and nondiagnostic CLE images (field of view = $475 \times 475$ \si{\micro\meter}). (a,b) Diagnostic images from glioma cases. (b) Unsupervisd localization of histopathological features of gliomas such as pleomorphism and hypercellularity detected by our model. For more results see Fig. \ref{localizationfig}. (c,d) Nondiagnostic images from meningioma cases occluded with motion (c) and blood artifact (d).}
\label{nondiagnostic}
\end{figure}

Current CLE systems are able to image more than one image per second, and thus over the course of examination of the surgical tumor resection or inspection area, hundreds to thousands of images may be collected. The number of images may become rapidly overwhelming for the neurosurgeon and neuropathologist when trying to quickly select a diagnostic or meaningful image or group of images as the surgical inspection progresses. CLE is designed to be used on the fly in real time while the surgeon operates the brain. Thus overcoming the barriers involved in image selection is a key component for making CLE a practical and advantageous technology for the neurosurgical operating room.

A wide range of fluorophores are able to be used for CLE in gastroenterology, but fluorophore options are limited for in vivo human brain use due to potential toxicity \cite{belykh2016intraoperative,foersch2012confocal,zehri2014neurosurgical}. In addition, motion and blood artifacts that are present in many of the images acquired with CLE using fluorescein sodium (FNa) are a barrier for revealing underlying meaningful histology. The display of suboptimal images or nondiagnostic frames interferes with the selection of and focus upon diagnostic images by the neurosurgeon and pathologist throughout the operation in order to make a correct intraoperative diagnosis. Previous assessment  \cite{martirosyan2016prospective} of CLE in human brain tumor surgery found that about half of the acquired images  were interpreted as nondiagnostic due to  abundance of motion and blood artifacts or lack of discernible or characteristic histopathological features.

Filtering out the nondiagnostic images before making an intraoperative diagnosis is challenging due to the high number of images acquired, the novel and frequently unfamiliar appearance of tissue features compared to conventional histology, great variability among images from the same tumor type, and potential similarity between images from other tumor types for the untrained interpreter (Fig. \ref{nondiagnostic}). 

Applications of machine learning in medical imaging have greatly increased in the last ten years, resulting in numerous computer-aided detection(CADe) and diagnosis(CADx) systems in ultrasound, magnetic resonance imaging (MRI), and computed tomography (CT) \cite{reviewDLMI}. Applications of machine learning for CLE imaging in neurosurgery are yet to be performed. In this study, we developed an ensemble of deep convolutional neural networks that can automatically evaluate the diagnostic value of CLE frames within milliseconds. 

Due to the limited number of images in our dataset, we sought to transfer learning benefits by using pretrained models, fine-tuning them in shallow and deep manner and compare results with the models trained from scratch. Our results demonstrated that a shallow fine-tuned model, although performs better than trained from scratch, is not enough for the optimal performance and that a deep fine-tuned model detects the diagnostic CLE frames better. We also investigated the effect of ensemble modeling by creating an ensemble of models which were crafted at the development stage and produced the minimum loss on validation dataset. Finally, we compared the performance among the ensemble of models and each single model.

\section{Related Works} 
\subsection{Convolutional Neural Networks}
Convolutional neural networks (CNNs), a subcategory of deep learning methods, have proven useful in visual imagery analysis from numerous fields, including medical images. This is mainly due to the deep multilayer architecture of CNNs which enables extracting abstract discriminant features, both local and global, present in the images.

In the recent years, deep learning has been vastly applied in medical image or exam classification. According to a survey done by Litjens et al. \cite{deepmedsurvey}, exam and object classification together make up the number one task of interest in medical image analysis followed by object detection and organ segmentation (exam classification alone is the third task of interest). Most of these studies in medical imaging field use one of the three following imaging modalities: MRI, microscopy or CT. 

Histopathological microscopic images and brain MRI scans were the first two areas where deep learning has been explored in medical imaging \cite{deepmedsurvey}. In histopathology, deep learning has been used for mitosis detection \cite{ciresan}, classification of leukocytes \cite{Zhao2016} and nuclei detection and classification \cite{sirinukunwattana}. In brain MRI, several studies have concluded that CNN benefits the diagnosis of Alzheimer’s disease \cite{suk2016deep, shi2017multimodal} as well as brain extraction \cite{salehi2017auto} and lesion detection, classification, and tumor grading \cite{ghafoorian2017deep,zhao2016multiscale}.

No-reference image quality assessment has been formulated as a classification problem as employed in retinal \cite{mahapatra2016retinal} and echocardiographic\cite{purang} images. CNNs may also be exercised in the detection of key frames from a temporal sequence of frames in a video. Two studies demonstrated the use of classification scheme on ultrasound (US) stream video to label the frames \cite{gao2016d,kumar2016plane}. 

\subsection{Transfer Learning vs. Deep Training}
One of the major limitations in medical imaging is the small size of datasets. The number of images employed for deep learning applications in medical imaging is usually much smaller than those in computer vision. Therefore, two forms of transfer learning have gained great interest: 1. Application of a pretrained network  on large-size natural images (i.e. ImageNet) as a feature extractor. 2. Initializing model parameters (weights and biases) using the data from a pretrained model \cite{yosinski2014transferable} instead of random initialization. A previous study by Tajbaksh et al. \cite{Nima} showed that a sufficiently fine-tuned AlexNet model could produce equal or better results than a deeply trained one for colonoscopy image quality assessment and few other medical applications. Here, we'll study the fine tuning effect by extending it to Inception network architectures in single and ensemble mode.

\subsection{Ensemble Modeling}
Ensemble modeling is a well established method for increasing the model performance and reducing its variance in machine learning \cite{dietterich2000ensemble,zhou2002ensembling,ciregan2012multi}. 
Kumar et al. \cite{kumar2017ensemble} created an ensemble of 5 different models to classify the image modality from ImageCLEF 2016 medical image dataset. Specifically, 2 classifiers were created by fine-tuning AlexNet and GoogLeNet with softmax and 3 other classifiers by training an SVM on top of the features extracted by AlexNet, GoogLeNet, and their combination. Their results showed the ensemble could improve the top-1 accuracy of the classifier compared to single models, however it is not clear that how much of the  improvement was because of the AlexNet and GoogLeNet combination or the 5 classifiers ensemble.  

Christodoulidis et al. \cite{christodoulidis2017multisource} created an ensemble of multi-source transfer learning using an automatic model selection approach described in [ensemble-selection]. After creating a pool of pretrained CNNs on several public texture datasets and fine-tuning them on the lung CT dataset, the top models which iterative grouping would produce the highest F-scores on the validation dataset were aggregated, creating an ensemble model. 5 ensemble models were developed and their output was then averaged to make the final output. Despite its computational complexity, it enhanced the lung disease pattern classification accuracy only by 2\%.

To generate diversity in our models while using the whole training dataset, we trained different neural networks on different data using cross-validation. Although previous studies have tried to create variant deep learning models using different network architectures, none of them have employed training data diversification through cross-validation as described in \cite{krogh1995neural}. Our proposed ensemble employed model diversification both in the network architectures and in the training and validation datasets following \cite{krogh1995neural}. 

\subsection{Confocal Laser Endomicroscopy in Neurosurgery}
Handheld, portable CLE has demonstrated its value for brain tumor surgery due to its ability to provide rapid intraoperative information regarding histopathological features of the tumor tissue \cite{martirosyan2016prospective}. Convenience, portability, and speed of CLE are significant advantages in surgery. A decision support system aiding and accelerating analysis of CLE images by the neuropathologist or neurosurgeon would improve the workflow in the neurosurgical operating room \cite{izady2017improv}.

Potentially used at any time during the surgery, CLE interrogation of the tissue generates images with a speed of 0.8 - 1.2 frames per second. The frames are considered nondiagnostic when the histological features are obscured by the red blood cells or motion artifacts, are out of focus, or lack any useful histopathological information. Acquired images are then exported from the instrument as JPEG or TIFF files for review. Currently, the pathologist reviews all images, including nondiagnostic ones, trying to explore the diagnostic frames for the diagnosis. Manual selection and review of thousands of images acquired at some point in surgery by the CLE operator is tedious and impractical for widespread use. Previously, we have presented \cite{izady2017improv} the first deeply trained CNN model for automatic detection of diagnostic CLE frames. In this work, we extend our previous work with the following contributions and advancements:

\begin{enumerate} 
 \setlength{\itemsep}{-2ex}  
 \setlength{\parskip}{0ex} 
 \setlength{\parsep}{0ex}
\item \textbf{Dataset}. Our dataset contains CLE images which is a novel technology in contrast to commonly used MRI or CT scans. The dataset used includes 20,734 CLE images from intracranial neoplasms.\hfil\linebreak
\item \textbf{Deep training, shallow fine-tuning or deep fine-tuning? } The CNN architectures were trained in three regimes: I. deeply trained (train the network from scratch with model weights randomly initialized) II. shallow fine-tuned (fine-tune only the fully connected layer(s) of the model which are responsible for the classification) III. deeply fine-tuned (fine-tune the whole network using our dataset).  In this study we report model accuracy on the test dataset for the best 5 models from each network architecture and training regime. Our work is different from \cite{Nima} since it considers the fine tuning effect on two different network architectures and its effect on the ensemble models.\hfil\break
\item\textbf{ Ensemble modeling}. Prior to the test phase, we created an ensemble of the best 5 models from each network and training regime. We explored the effect of ensemble modeling in all circumstances by comparing the ensemble performance with the average of single models. Our work is different from \cite{kumar2017ensemble} since our ensemble generates diversity in single models by using different training and validation data achieved from nested-cross validation. Further, we studied the effect of ensemble modeling on different training scenarios rather than one.  \hfil\break
\item \textbf{Unsupervised localizing of histological features}. We visualized the shallow neurons' activation to depict the broad histological patterns; visualization of deep neurons’ activation could localize specific histopathological lesions for diagnostic images. The neural response of convolutional layers to the diagnostic images are visualized and analyzed by a neurosurgeon. We also extracted the CNN's deepest neural activation in response to patches of the diagnostic images using a sliding-window. Qualitative assessment of the localized regions was performed by a neurosurgeon with further analysis of the histopathological features.\hfil\break
\item \textbf{Interobserver study.} We compared the interobserver agreement between physician-physician and ensemble of models-physician to compare our ensemble model performance with human performance. We also reported the kappa statistic for this observer study.\hfil\break

\end{enumerate}


\section{Methods}

\subsection{Image Acquisition}
In the following sections we briefly explain the confocal imaging instrument instrument specifications and the intraoperative data collection process.
\subsubsection{Instrument specifications}

The CLE image acquisition system (Optiscan 5.1, Optiscan Pty, Ltd.) included a rigid pen-sized optical laser scanner with a 6.3 \si{\milli\meter} outer diameter and a working length
of 150 \si{\milli\meter}. A 488 \si{\nano\meter} diode laser provided 
excitation light and the fluorescent emission signal was detected with a ~505-585 \si{\nano\meter} band-pass filter. A single optical fiber
acted as both the excitation pinhole and the detection pinhole for confocal isolation of the image plane. The detector signal was digitized synchronously with the scanning to construct images parallel to the tissue surface (en face optical sections).

Laser power was typically set to 550-900 \si{\micro\watt} and maximum power was limited to 1000 \si{\micro\watt} when applied to the brain tissue. A field of view of $475 \times 475$ \si{\micro\meter} was scanned at
$1024 \times 1024$ pixels (1.2/second frame rate), with a lateral
resolution of 0.7 \si{\micro\meter} and an axial resolution (i.e., effective
optical slice thickness) of approximately 4.5 \si{\micro\meter}.
\subsubsection{Intraoperative CLE imaging}
Seventy-four adult patients (31 male and 43 female) were enrolled in the study (mean age 47.5 years). Intraoperative CLE images were acquired both in vivo and ex vivo by 4 neurosurgeons. For in vivo imaging, multiple locations of the tissue around the lesion were imaged and excised from the patient. For ex vivo imaging, tissue samples suspicious for tumor were excised, placed on gauze and imaged on a separate work station in the operating room. Multiple images were obtained from each biopsy location.

Co-registration of the CLE probe with the image guided surgical system allowed precise intraoperative mapping of CLE images with regard to the site of the biopsy. The only fluorophore administered was FNa (5\si{\milli\liter}, 10\%) that was injected intravenously during the surgery.

Precise location of the areas imaged with the CLE was marked with tissue ink. Imaged tissue was sent to the pathology laboratory for formalin fixation, paraffin embedding and histological sections preparation. Final histopathological assessment was performed by standard light microscopic evaluation of 10-\si{\micro\meter}-thick hematoxylin and eosin (H \& E)-stained sections. 

\subsection{Image annotation}
The image annotation process was done in two distinct stages: \textit{initial review} and \textit{
}.
\subsubsection{Initial review}\label{review1} Initially all  images were reviewed. A neuropathologist and 2 neurosurgeons who were not involved in the surgeries reviewed the CLE images. For each patient, the histopathological features of corresponding CLE images and H \& E-stained frozen and permanent sections were reviewed and the diagnostic value of each image was examined. When CLE image revealed clear identifiable histopathological feature, it was labeled as \textit{diagnostic}; otherwise it was labeled as \textit{nondiagnostic}. 
\subsubsection{Validation review}\label{valreveiw}
 The database of images was divided into development and test datasets (explained in dataset preparation \ref{Data Preparation}). Test dataset composed of 4171 CLE images randomly chosen from various patients. The \textit{validation review} (val review) dataset consists of 540 images randomly chosen from the test dataset. Following this separation, two neurosurgeons reviewed val-review dataset without having access to the corresponding H \& E-stained slides and labeled them as diagnostic or nondiagnostic . 

\subsection{Convolutional Neural Networks}

Convolutional Neural Networks  (CNNs) are multilayer learning frameworks and may consist of an input layer, a few convolutional layers, pooling layers, fully connected layers and the output. The goal of a CNN is to learn the hierarchy of underlying feature representations. We explain the fundamental elements of a CNN below.

\subsubsection{Convolutional layer}
Convolutional layers, first introduced in \cite{LeCun:1998:CNI:303568.303704} are the substitute of previous hand-crafted feature extractors. At each convolutional layer three dimensional matrices (kernels) are slid over the input and set the dot product of kernel weights with the receptive field of the input as the corresponding local output. This helps to retain the relative position of features to each other. The multi-kernel characteristic of convolutional layers enables them to prospectively extract several distinct feature maps from the same input image.
\subsubsection{Activation layer}

The convolutional layer output then goes through an activation function to adjust the negative values. We employed the rectified linear unit (ReLU) which is usually the preferred choice because of its simplicity, higher speed, reduced likelihood of vanishing gradients (especially in deep networks) and tendency to add sparsity over other nonlinear functions such as sigmoid function. The output of $j^{th}$ ReLU layer ($a_j^{out}$), given its input ($a_j^{in}$), was calculated in-place (to consume less memory) by following:
\begin{equation} \label{relu}
a_j^{out} = \max{(a_j^{in},0)}
\end{equation}
\subsubsection{Normalization layer}

Following the ReLU layer, a local response normalization (LRN) map is applied after the initial convolutional layers. This layer inhibits the local ReLU neurons' activations since there's no bound to limit them (Eq. \ref{relu}). By using the Caffe implemented \cite{jia2014caffe} LRN, the local regions are expanded across neighbor feature maps at each spatial location. The output of $j^{th}$ LRN layer ($a_j^{out}$), given its input ($a_j^{in}$), is calculated as:
\begin{equation}
a_j^{out} = \frac{a_j^{in}}{(1+\frac{\alpha}{L}\sum_{n=1}^L {{a_j^{in}(n)}^2})^\beta}
\end{equation}

where ${a_j^{in}(n)}$ is the $n^{th}$ element of the $a_j^{in}$ and $L$ is the length of $a_j^{in}$ vector (number of neighbor maps employed in the normalization). $\alpha$, $\beta$ and $L$ are the layer's hyperparameters and are set to their default values obtained from \cite{krizhevsky2012imagenet}($\alpha=1$, $\beta=0.75$ and $L=5$).

\subsubsection{Pooling layer}

After rectification and normalization of convolutional layer output, it's further down-sampled by pooling operations. Pooling operations accumulate values in a smaller region by subsampling operations such as max, min, and average sampling. Here, max pooling was applied.

\subsubsection{Fully connected Layer}
Following several convolutional and pooling layers, the network lateral layers are fully connected. Each neuron of the layer's output is greedily connected to all the layer's input neurons. It can be thought of as a convolutional layer with kernel size of the layer input. The layer output is also passed through a ReLU layer. 
The fully connected layers are generally thought of as the classifier of a CNN model because they intake the most abstract features extracted in convolutional layers and make the output, which is the model prediction. 

\subsubsection{Dropout Layer}
Fully connected layers are usually followed by a dropout layer, except the last fully connected layer that produces the class-specific probabilities. In dropout layers, a subset of input neurons as well as all their connections are temporarily removed from the network. Srivastava et al.\cite{dropout} have demonstrated this method efficiency at improving the CNN performance in numerous computer vision tasks through reducing the overfitting.

\subsubsection{Learning}
The learning of a CNN is through Stochastic Gradient Descent (SGD) which stands on two major menhirs: Forward and Back Propagation. In forward propagation, the model makes predictions using the images in the training batch and the current model parameters. Once the prediction for all training images is made, the loss is calculated using the truth label provided by the experts in the initial review (explained in \ref{review1}). In this work we adopt the softmax loss function given by:
\begin{equation}
L(t,y) = -\frac{1}{N}\sum_{n=1}^N \sum_{k=1}^C t_k^n log (\frac{e^{y_k^n}}{\sum_{m=1}^C e^{y_m^n}})
\end{equation}
where $t_k^n$ is the $n^{th}$ training image's $k^{th}$ ground truth output, and $y_k^n$ is the value of the $k^{th}$ output layer unit in response to the $n^{th}$ input training image. $N$ is the number of training images in the minibatch, and since we consider $2$ diagnostic value categories, $C=2$. 

Through the back propagation, the loss gradient with respect to all model weights aids upgrading the weights as follows:
\begin{equation}
W(j,i+1) = W(j,i) + \mu \Delta{W(j,i)} - \alpha(j,i) \frac{\partial L}{\partial W(j)}
\end{equation}
where $W(j,i)$, $W(j,i+1)$ and $\Delta{W(j,i)}$ are the weights of $j^{th}$ convolutional layer at iteration $i$  and  $i+1$ and the weight update of iteration $i$, $\mu$ is the momentum and $\alpha(j,i)$ is the learning rate and is dynamically lowered as the training progresses.

\subsection{Evaluation Metrics}
In model performance estimation (explained in \ref{testing}) we calculated the \textit{loss, accuracy, sensitivity, specificity and area under receiver operating characteristics (ROC) curve (AUC)}. Here, we assumed the state of being a diagnostic image as positive and being nondiagnostic as negative. This way, \textit{sensitivity} determines the model ability to detect diagnostic images and \textit{specificity} determines its ability to detect nondiagnostic images. \textit{Accuracy} determines  general capability of a model to  detect diagnostic and nondiagnostic images correctly \cite{metz1978basic}.


\section{Experimental Setup}

\subsection{Dataset Preparation} \label{Data Preparation}
Our dataset included 20,734 CLE images from 74 brain tumor cases. For each CLE image, the diagnostic quality was determined by the experts in the initial review. The dataset was divided into two main subsets on patient level: \textit{development (dev)} and \textit{test}. Our pilot study revealed that when the division is on image level (mixing the images from all the patients and dividing them randomly) the model would produce poor results on images from new patients.

The total number of patients and images used at each stage are presented in Table \ref{dataprep}. Each subset contains images from various tumor types (mainly from gliomas and meningiomas). The dev set will be available online. The test set was isolated all through the model development and was accessed only in the test phase.

\begin{table}[hb!]
\centering
\caption{Dataset preparation: Patient-based allocation of diagnostic and nondiagnostic images from various neoplasms to model development and testing. Number of patients for each tumor type is also provided.}
\label{dataprep}
\begin{tabular}{@{}cccc@{}}
\toprule
\textbf{}                     & \textbf{}                                    & \textbf{Development}    & \textbf{Test} \\ \midrule
\rowcolor[HTML]{C0C0C0} 
\multicolumn{2}{l}{\cellcolor[HTML]{C0C0C0}\textbf{No. of Patients (total)}} & \textbf{59}     & \textbf{15}   \\
\textbf{}                     & \textbf{Gliomas}                             & \textbf{16}     & \textbf{5}    \\
\textbf{}                     & \textbf{Meningiomas}                         & \textbf{24}     & \textbf{6}    \\
\textbf{}                     & \textbf{Other neoplasms}                     & \textbf{19}     & \textbf{4}    \\
\rowcolor[HTML]{C0C0C0} 
\multicolumn{2}{l}{\cellcolor[HTML]{C0C0C0}\textbf{No. of Images (total)}}   & \textbf{16,366} & \textbf{4,171} \\
\textbf{}                     & \textbf{Diagnostic}                          & \textbf{8,023}   & \textbf{2,071} \\
\textbf{}                     & \textbf{Nondiagnostic}                       & \textbf{8,343}   & \textbf{2,100} \\ \bottomrule
\end{tabular}
\end{table}



\begin{figure*}[th!]
\centering
\includegraphics[trim={7cm 0 5cm 0},clip,width=4.8in]{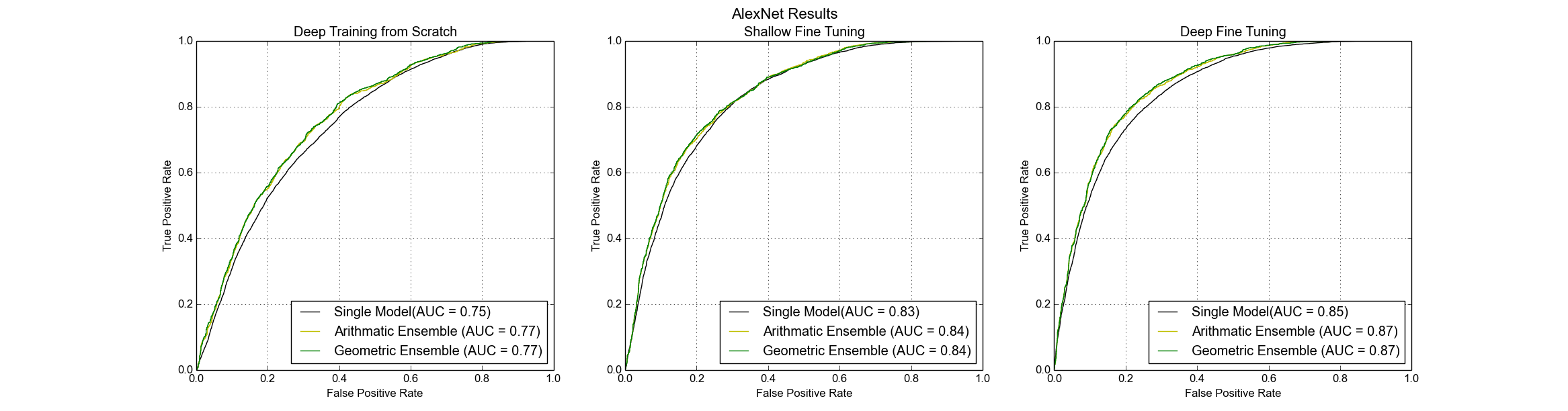}\\
\includegraphics[trim={7cm 0 5cm 0},clip,width=4.8in]{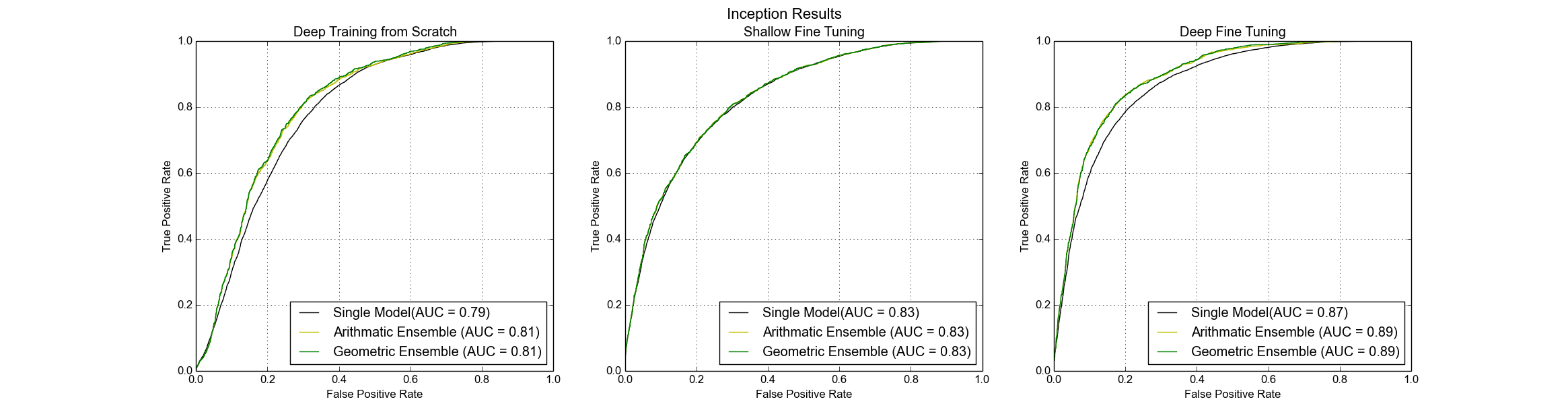}
\caption{Ensemble effect on network 1 (top) and network 2 (bottom) while using diverse training regimes. For both networks, the improvement was more noticeable with DT and DFT regimes. The arithmetic and geometric ensemble  performed similarly. Neither of the two ensembles could improve network 2 trained with SFT.}
\label{Ens}
\end{figure*}

\subsection{Model Development}
After the initial data split, we employed a patient-based k-fold cross validation for model development. Fifty nine cases that were allocated for model development were divided into 5 groups. Since CNNs require a large set of hyperparameters to be defined optimally (i.e. initial value of the learning rate and its lowering policy, momentum, batch size, etc.), we used different values with grid searching throughout the model development process. For every set of feasible parameters, we trained the model on 4 folds and validated on the fifth left-out group of patients. The set of hyperparameters which produced the minimum average loss was employed for each set of experiments.

The small dataset size was a main limitation of our study for using CNNs, especially with the patient-level data preparation. Therefore, we counterbalanced this limitation by fine-tuning the pretrained  publicly available CNN architectures trained on large computer vision datasets (i.e. ImageNet). 

Though the question about \textit{how deep should we fine-tune the pretrained models for optimal results} still remains unanswered, one study tried to answer this question using endoscopy and ultrasound images\cite{Nima}. Due to the substantial intrinsic dissimilarities between the images in the 2 studies, we performed a similar investigation. Our confocal images have a much higher spatial resolution and are fluorescent images from the brain. 

In total, we developed 42 models (30 single models and 12 ensemble models) using two network architectures and three training regimes (deep training, shallow fine-tuning and deep fine-tuning). The experiments are designed in order to practically find the optimal model development pathway that produces the highest performance in the considered clinical application. 

\subsubsection{Network architectures}
Two CNN architectures were applied in this study. 
Network 1 had 5 convolutional layers. The first two convolutional layers had $96$ and 256 filters of size $11\times11$ and $5\times5$ with maximum pooling. The third, fourth and fifth convolutional layers were connected back to back without any pooling in between. The third convolutional layer had $384$ filters of size $3\times3\times256 $, the fourth layer had $384$ filters of size $3\times3\times192 $ and the fifth layer had $256$ filters of size $3\times3\times192 $ with maximum pooling. For more details please refer to \cite{krizhevsky2012imagenet}.

Network 2 had 22 layers with parameters and 9 inception modules. Each inception module was a combination of filters of size $1\times1$, $3\times3$, $5\times5$ and a  $3\times3$ max pooling, put together in parallel and the output filter banks concatenated into an input single vector for the next stage. For more details please refer to \cite{szegedy2015going}.

The pretrained model for network 1, exploited in fine-tuning experiments, was the iteration 360,000 snapshot of training the model on ImageNet classification with 1000 classes. The pretrained model for network 2 was iteration 2,400,000 of training on ImageNet classification dataset. Both models are publicly available in Caffe libraries \cite{jia2014caffe}. 

\subsubsection{Training regimes}
We exercised various training regimes to see how deep fine-tuning should be done in CLE image classification for optimal results. Depending on which layers of the network are being learned through training, we had three regimes.

In regime 1, \textit{\textbf{deep training (DT)}}, the whole model weights were initialized randomly (training from scratch) and got modified all through the training with nonzero learning rates. 

In regime 2, \textit{\textbf{shallow fine-tuning (SFT)}}, the whole model weights, except the last fully connected layer, were initialized with the corresponding values from the pretrained model and their values were fixed for the period of training. The last fully connected layer was initialized randomly and got tuned during training.

In regime 3, \textit{\textbf{deep fine-tuning (DFT)}}, all model weights, except for the last fully connected layer, were initialized with the corresponding values from the pretrained model and last fully connected layer was initialized randomly. Throughout the training, all the CNN layers, including the last fully connected layer, were tuned with nonzero learning rates. Our hyperparameter optimization showed that the SFT and DFT experiments required 10 times smaller initial learning rates (0.001) compared to the DT regime (0.01).
To avoid overfitting, the training process was stopped after 3 epochs of consistent loss increment on the validation dataset. We also used dropout layer ($ratio=0.5$) and $L2$ regularization ($\lambda=0.005$).

\subsubsection{Ensemble Modeling}
Let's assume $y_k^n(j)$ is the the value of the $k^{th}$ output layer unit of the $j^{th}$ CNN model in response to the $n^{th}$ input test image. The linear and log-linear ensemble classifier output for the same input would be:

\begin{equation}\label{lin}
Ens_{linear}^n = \text{arg}\,\max\limits_{k} \sum_{j=1}^l{y_k^n(j)}
\end{equation}

\begin{equation}\label{loglin}
Ens_{log-linear}^n = \text{arg}\,\max\limits_{k} \prod_{j=1}^l{y_k^n(j)}
\end{equation}

where l is the number of CNN models combined to generate the ensemble models.

Model selection was done in two forms: single models and ensemble of models. We selected the top model (with minimum loss on the validation dataset) from each fold of the 5-fold cross validation (\textbf{Model 1-5} in Table \ref{results_tab}). In each network architecture and training regime, we combined the top-5 developed single models to produce two ensembles of models using the arithmetic (\ref{lin}) and geometric mean (\ref{loglin}) of their outputs. We created 12 ensemble models ($2^{network\, architectures} \times 3^{training\, regimes} \times 2^{ensemble\, types}$) in total and compared their performance with single models.

\subsection{Interobserver Study} \label{testing}
 Each solo and ensemble model developed was tested on the test dataset. The ensemble of network 2 models trained with DFT was also tested on the \textit{val review } images (\ref{valreveiw}) to compare human-human and model-human interobserver agreements. The resulting agreement rate (val-rater 1 and 2) was further compared with the initial image review results. The agreement of the model prediction with the initial review was also calculated. The general agreements are compared and discussed in section \ref{Inter-rater}. Kappa analysis was also done for further validation.

\begin{figure}[ht!]
 \includegraphics[width = 4.8in]{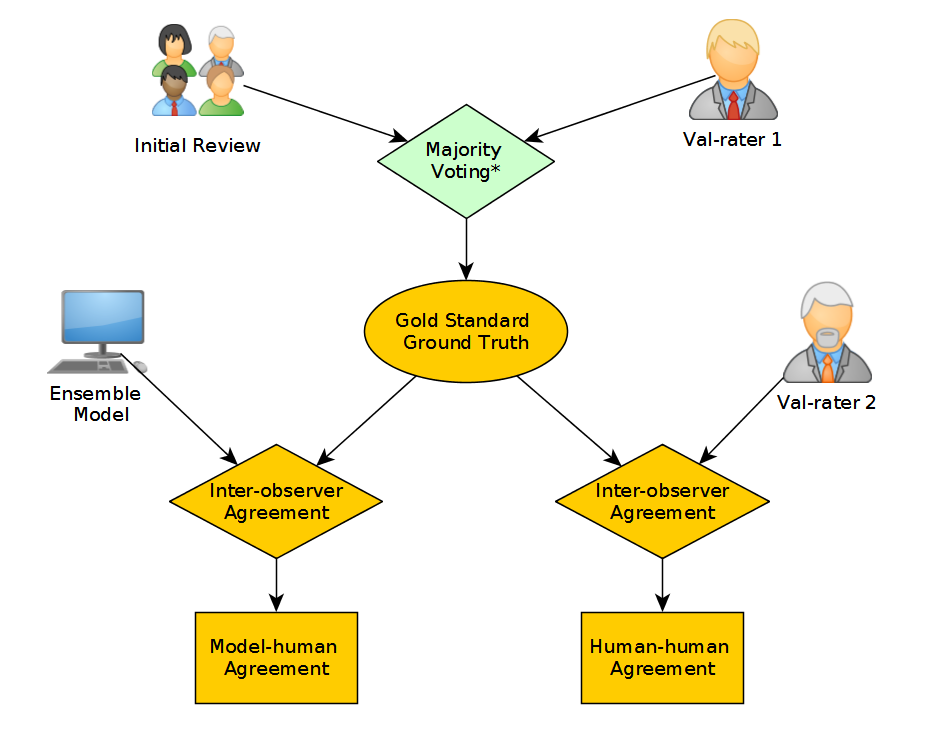}
\centering
 \caption{Interobserver study using gold standard ground truth. Gold standard was defined using the initial review and one of the val-raters (here val-rater 1). Then, the agreement of the ensemble model and the other val-rater (here val-rater 2) with the gold standard is calculated to compare the human-human with model-human agreement.
*If the initial review and the val-rater 1 agreed on an image, it is added to the gold standard, otherwise it is disregarded.}
 \label{IOfig}
 \end{figure}

 Gold standard ground truth for the val review images  was defined by majority voting (see Fig. \ref{IOfig}). The agreements of the third rater with the gold standard and the proposed ensemble model with the gold standard is calculated and compared as well in Table \ref{interrater results}.

\subsection{Unsupervised Histological Feature Localization}
For localization of the histological features, we examined the neural activation at two sites. First, the activation of neurons in the first convolutional layer of the network 1 were visualized and the 96 feature planes were saved for review by a neurosurgeon. Neurons that presented high activation to the location of cellular structures in the input image were selected and seemed to be consistent with diverse diagnostic images. Secondly, we applied a sliding window of size $227\times227$ pixels (size of network 1 input after input cropping) with stride of 79 pixels over the diagnostic CLE images ($1024\times1024$ pixels). The result was a $10\times10$ matrix that provided the diagnostic value of different locations of the input image (\textit{diagnostic map}). The locations of input images corresponding to the highest activations of the diagnostic map were detected and marked with a bounding box. The detected features using each of these two ways were further reviewed by a neurosurgeon.
\section{Results and Discussion}\label{sectionresults}
 We developed 42 models and tested them on 4,171 test images; accuracy rates (agreement with the initial review) are presented in Table \ref{results_tab}. We found that network 2 resulted in more precise predictions about the diagnostic quality of images than network 1 when  DT and DFT training regimes were used, while  SFT training regime resulted in slightly better accuracy of network 1, compared to network 2. Therefore,  network 2 architecture is a better feature extraction tool for CLE images, since it concatenates multi-scale features inside its inception modules.

\begin{table}[ht!]
\centering
\caption{The accuracy of different models on the test dataset. The top-5 models crafted from each training regime, as well as their arithmetic and geometric ensembles, were exerted. For each network, the ensemble of DFT models makes the most accurate predictions. The difference between arithmetic and geometric ensemble AUC was negligible.}
\label{results_tab}
\begin{tabular}{|c|c|c|c|c|c|c|}
\hline
\textbf{Network}                                                       & \multicolumn{3}{c|}{\textbf{Network 1}}            & \multicolumn{3}{c|}{\textbf{Network 2}}          \\ \hline
\textbf{\begin{tabular}[c]{@{}c@{}}Training\\ Regime\end{tabular}}     & \textbf{DT}    & \textbf{SFT}   & \textbf{DFT}   & \textbf{DT}    & \textbf{SFT}   & \textbf{DFT}   \\ \hline
\textbf{Model 1}                                                       & 0.685          & 0.760          & 0.760          & 0.731          & 0.746          & 0.746          \\ \hline
\textbf{Model 2}                                                       & 0.658          & 0.749          & 0.755          & 0.750          & 0.746          & 0.805          \\ \hline
\textbf{Model 3}                                                       & 0.677          & 0.751          & 0.765          & 0.715          & 0.747          & 0.797          \\ \hline
\textbf{Model 4}                                                       & 0.681          & 0.754          & 0.771          & 0.739          & 0.743          & 0.811          \\ \hline
\textbf{Model 5}                                                       & 0.699          & 0.753          & 0.775          & 0.721          & 0.747          & 0.777          \\ \hline
\textbf{Mean}                                                          & 0.680          & 0.753          & 0.765          & 0.731          & 0.746          & 0.787          \\ \hline
\textbf{\begin{tabular}[c]{@{}c@{}}Arithmatic\\ Ensemble\end{tabular}} & \textbf{0.704} & 0.755          & \textbf{0.788} & 0.754          & 0.750          & 0.816          \\ \hline
\textbf{\begin{tabular}[c]{@{}c@{}}Geometric\\ Ensemble\end{tabular}}  & 0.703          & \textbf{0.758} & 0.786          & \textbf{0.755} & \textbf{0.751} & \textbf{0.818} \\ \hline
\end{tabular}
\end{table}

\begin{figure*}[ht!]
\centering
\includegraphics[trim={7cm 0 5cm 0},clip,width=4.8in]{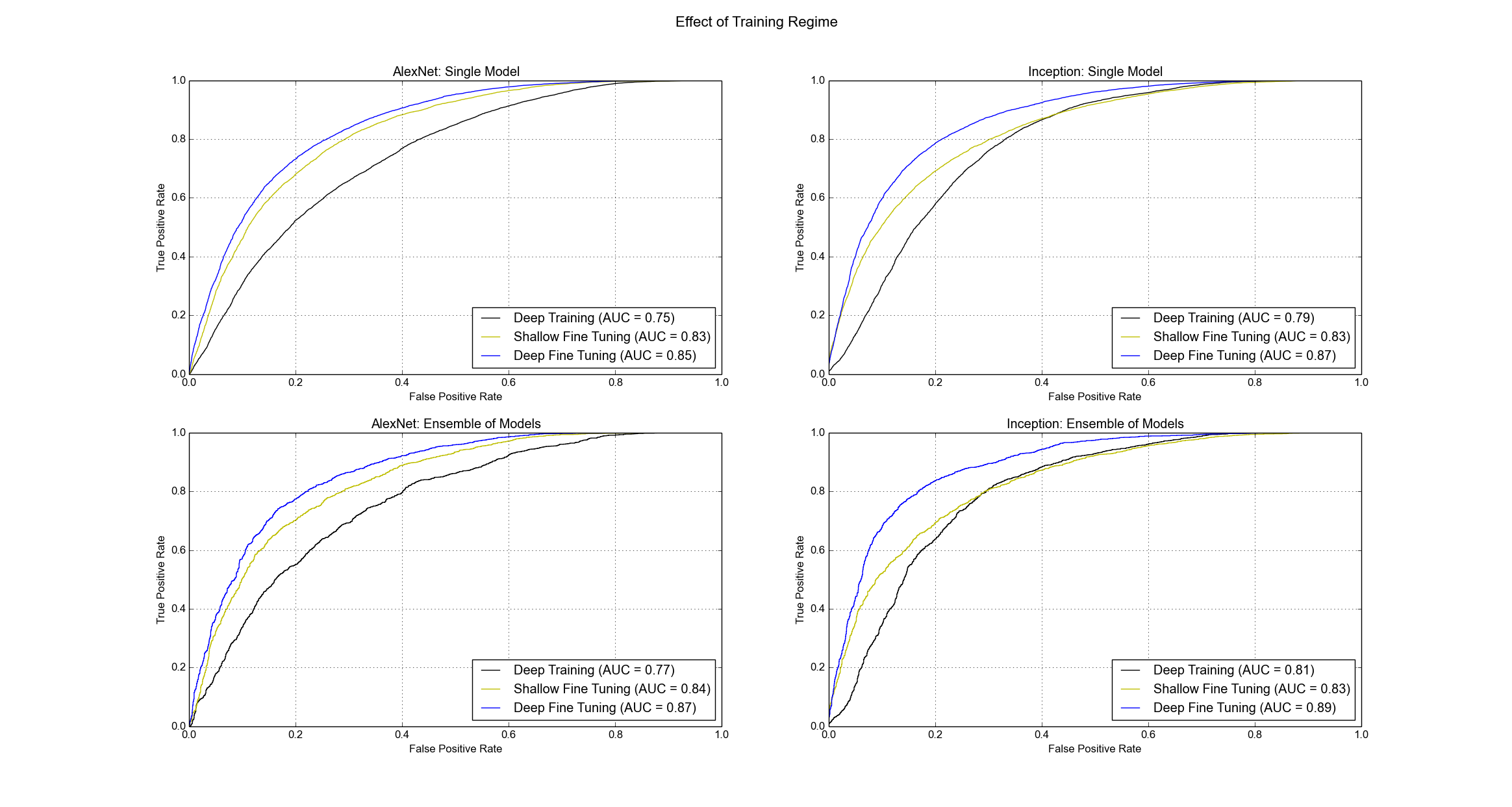}
\caption{Training regime effect on network 1 (left) and network 2 (right) while using single (top) or ensemble of models (bottom). In all circumstances the AUC for DFT regime was greater than the SFT and SFT is greater than DT, although the effect size varied.}
\label{regimes}
\end{figure*}

\begin{figure*}
\centering
\subfloat[]{\includegraphics[width = 1.15in]{486.jpg}} ~
\subfloat[]{\includegraphics[width = 1.15in]{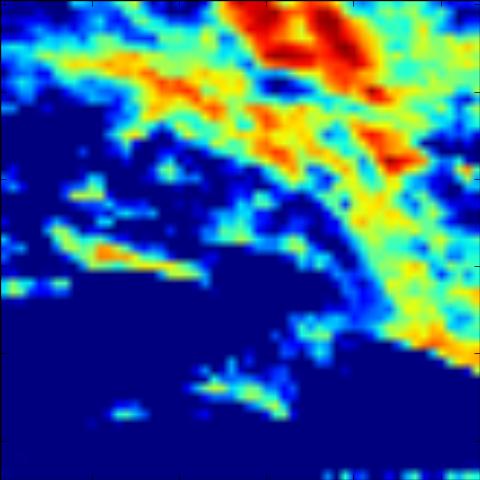}} ~
\subfloat[]{\includegraphics[width = 1.15in]{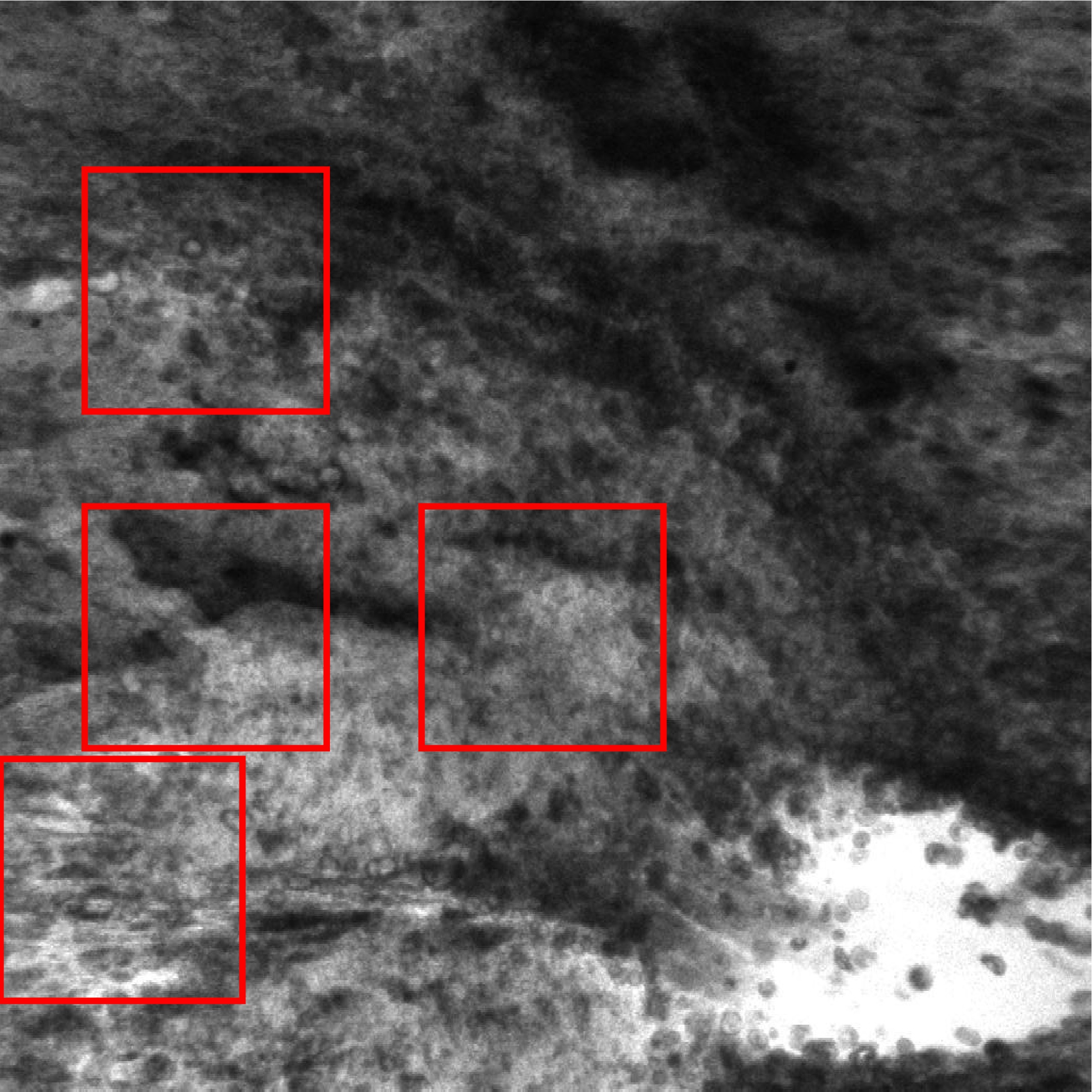}} ~
\subfloat[]{\includegraphics[width = 1.15in]{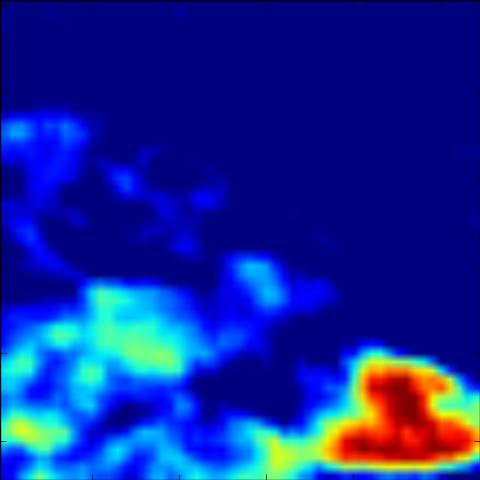}}
\\%
\subfloat[]{\includegraphics[width = 1.15in]{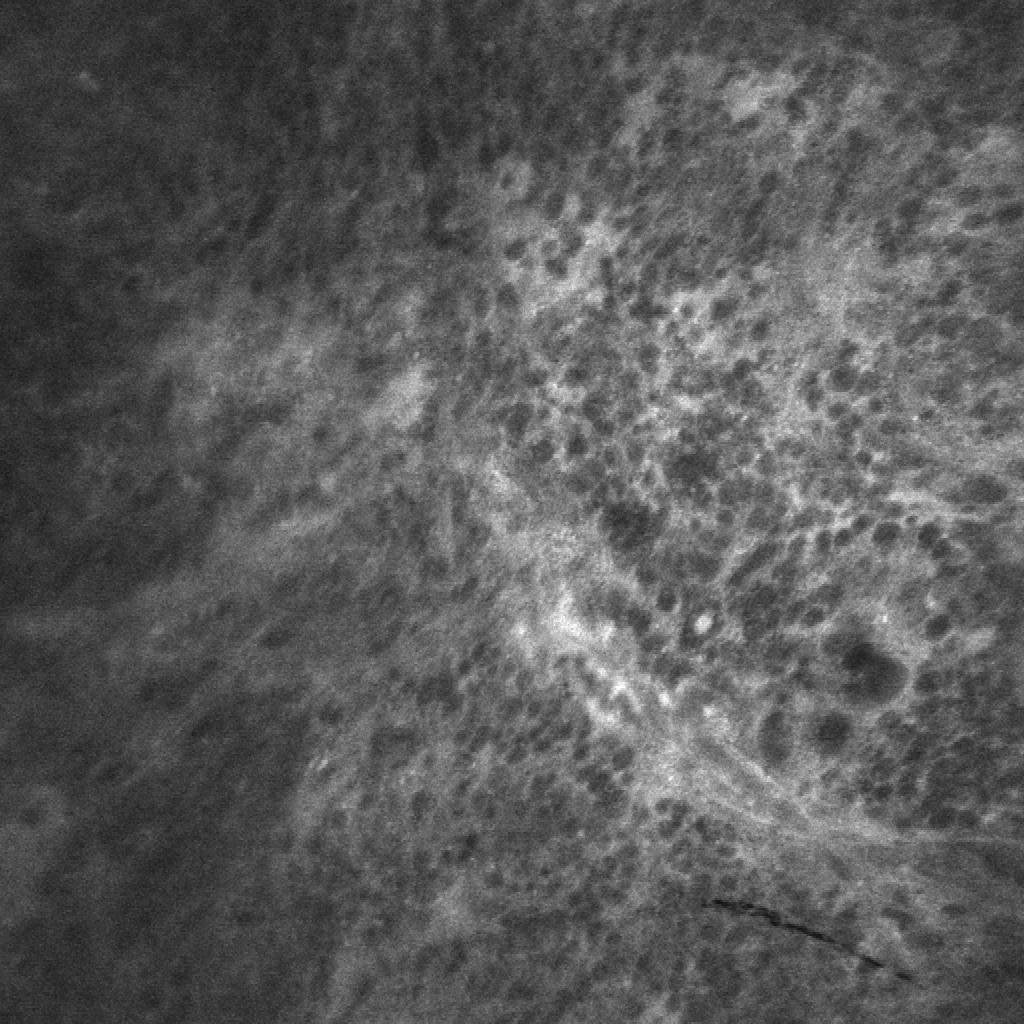}} ~
\subfloat[]{\includegraphics[width = 1.15in]{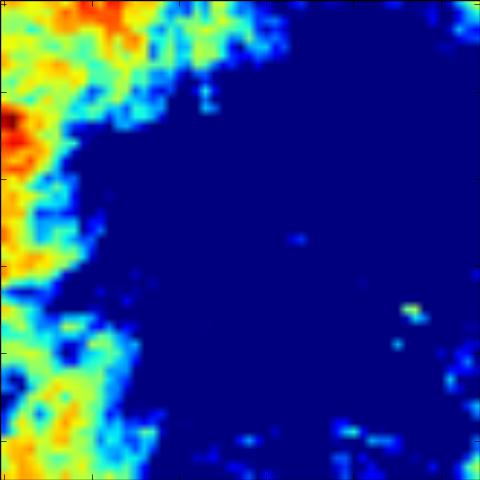}} ~ 
\subfloat[]{\includegraphics[width = 1.15in]{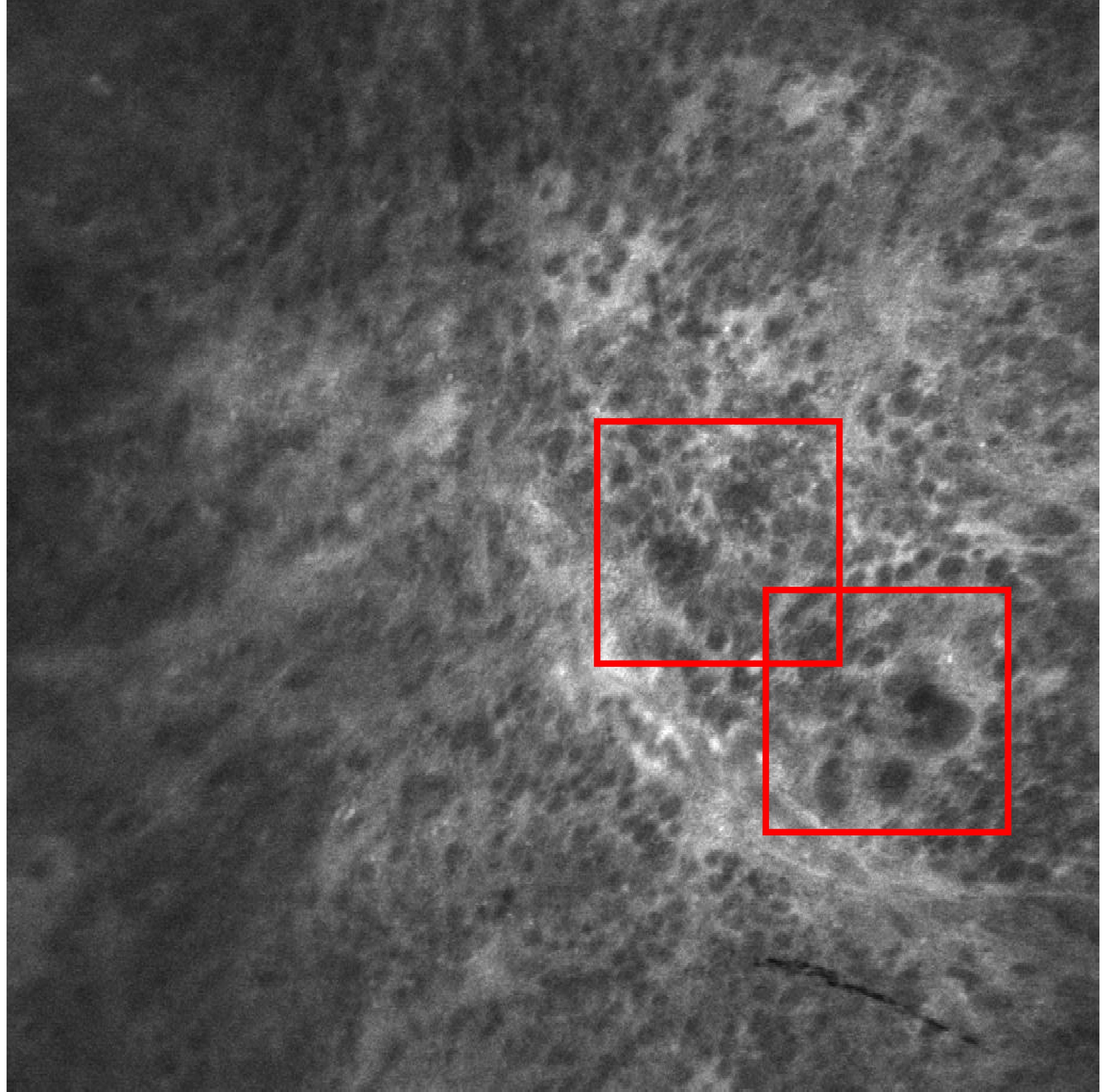}} ~
\subfloat[]{\includegraphics[width = 1.15in]{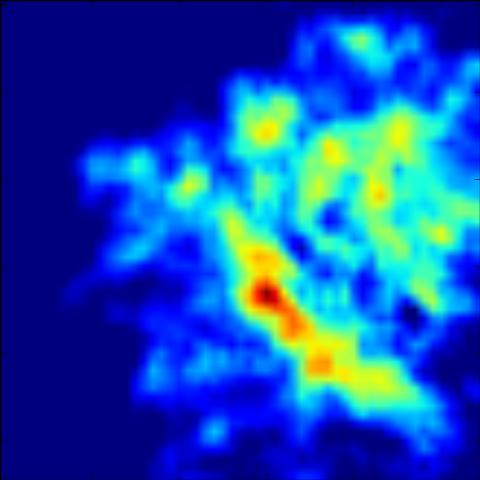}}
\\%
\subfloat[]{\includegraphics[width = 1.15in]{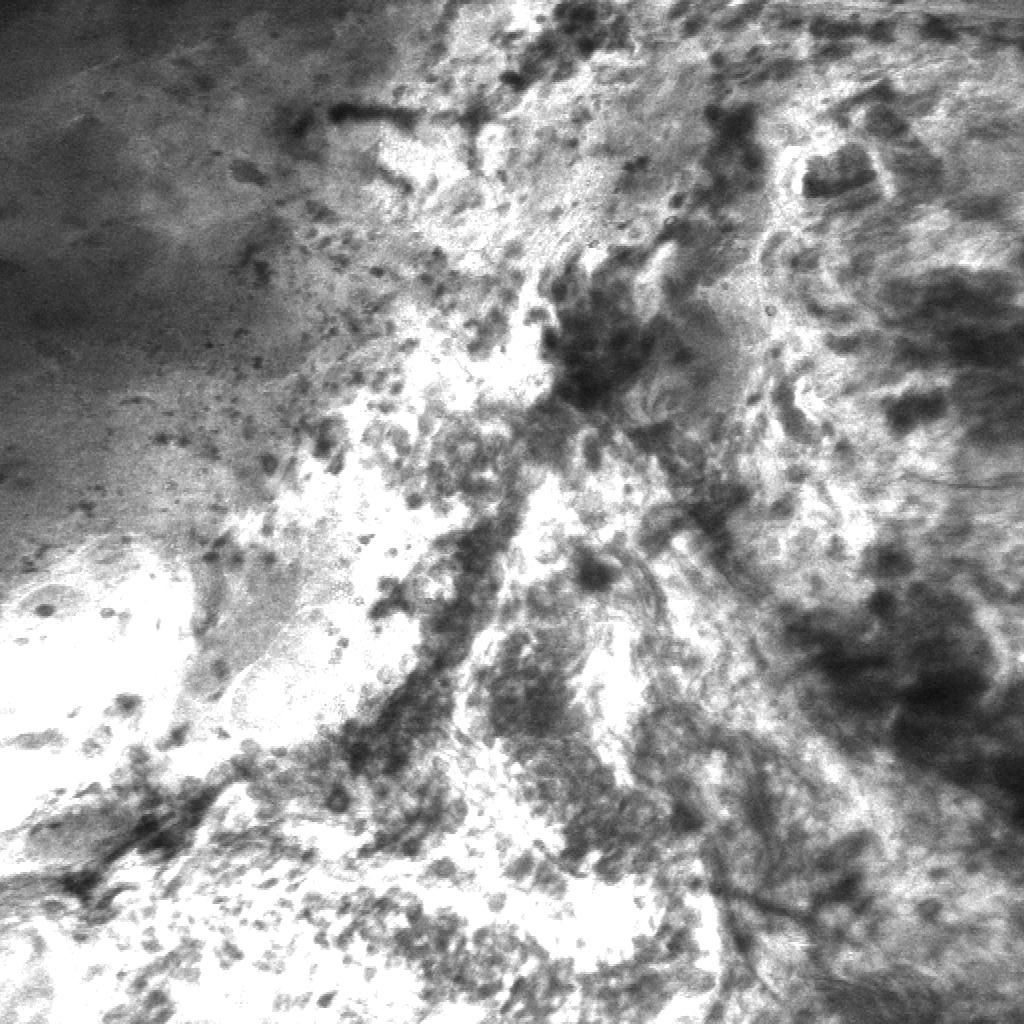}} ~
\subfloat[]{\includegraphics[width = 1.15in]{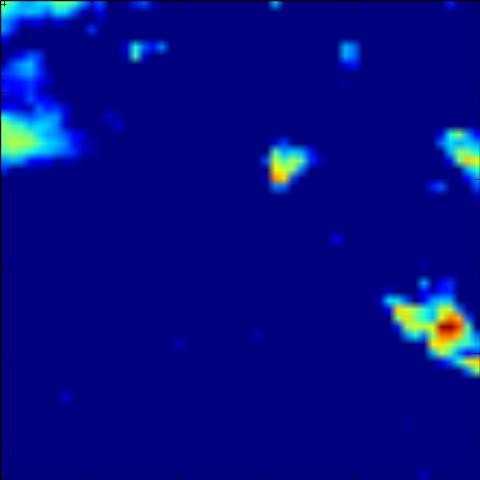}} ~
\subfloat[]{\includegraphics[width = 1.15in]{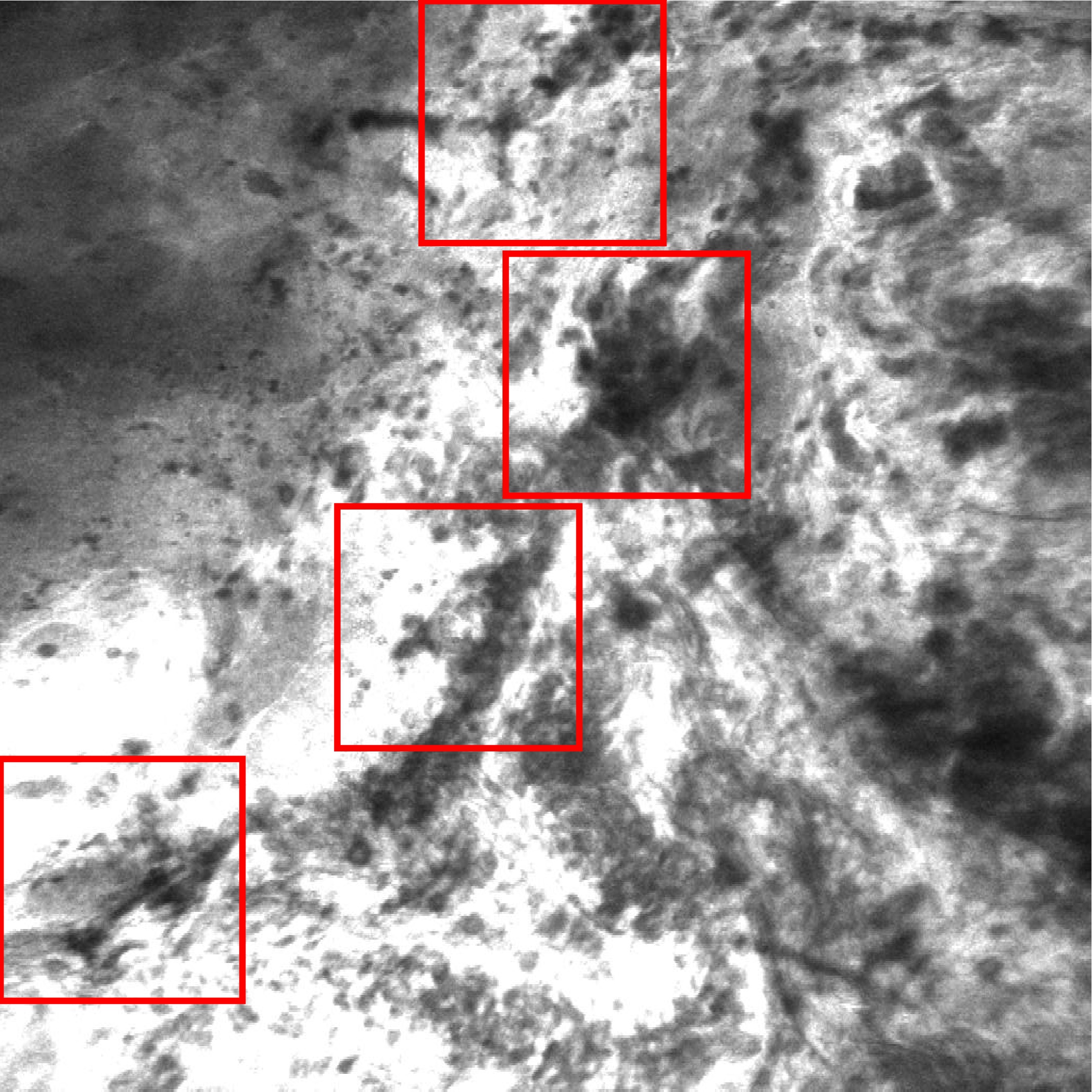}} ~
\subfloat[]{\includegraphics[width = 1.15in]{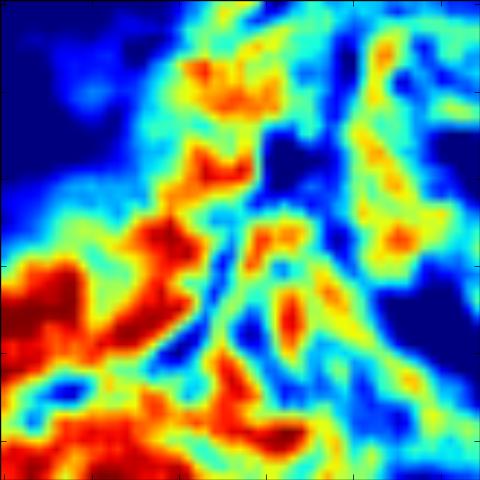}}
\\%
\caption{Unsupervised localization of the histopathological features from shallow and deep neurons inside the network. First  column (a, e, i) shows the input CLE images from human glioblastoma obtained intraoperatively. Second column (b, f, j) displays activation of neurons from the first layer (conv1, neuron 24) (shallow features); it highlights some of the cellular areas present in the image. Third column (c, g, k) illustrates diagnostic regions of interest identified with sliding window approach. The boxed regions represent high activation of the deepest network neuron. Fourth  column (d, h, l) contains images extracted from conv1 activation (neuron
22), representative of the high fluorescence signal, a diagnostic sign of blood-brain barrier disruption and leakage of fluorescent agent from the vessels into the extracellular space.} 
\label{localizationfig}
\end{figure*}

\subsection{Ensemble or solo model?}
We did an ROC analysis for each of the two networks and three training regimes to see how the ensemble of models performed compared to the single models. Fig. \ref{Ens} presents the ROC curves and corresponding AUC values for each ensemble model and the mean of single models. The AUC value increased by 2\%  for both networks with DT and DFT when the ensemble is applied instead of the single model. This effect gets smaller with network 1 SFT and becomes negligible with network 2 SFT. The two arithmetic and geometric ensemble models produced roughly similar results (paired t-test: \textit{P value $<$ 0.05}). 

SFT models display less sensitivity to the ensemble effect compared to DT and DFT. This is not surprising since they represent identical models except in the softmax classifier layer which has been adjusted through training.

\subsection{Which training regime: DT, SFT and DFT?}
Fig. \ref{regimes} displays the results of ROC analysis when comparing the three training regimes in each network architecture and single/ensemble states. In all paired comparisons, DFT outperformed  SFT and SFT outperformed the DT regime (paired t-test: \textit{P value $<$ 0.05}). 

We traced the AUC elevation from DT to DFT regime to see how much of it corresponded to the transformation of DT to SFT and SFT to DFT. For network 1, 70-80 \% of the improvement occurred in the DT to SFT transformation, depending on whether it's a single or ensemble model. For network 2 ensemble model (right bottom of Fig. \ref{regimes}), however, the AUC improvement caused by transforming the training regime from DT to SFT (2\%) is only 25\% of the total improvement from DT to DFT. For network 2 single model the AUC improvement was evenly divided between the two transformations.

Our results from this experiment indicated that for our dataset, network 1 mainly required fine-tuning the classification layer and fine-tuning other layers (feature extractors) had a smaller contribution. However, for network 2, fine-tuning the feature extractors was more important than modifying the classifier layer. Though, further experiments on more datasets are necessary to generalize this observation.
\subsection{Histological features localization}
 8 out of total 384 reviewed colored neuron activation maps from the first layer were selected for 4 diagnostic CLE images representative for glioma. Selected activation maps highlighted diagnostic tissue architecture patterns in warm colors. Particularly, several maps emphasized regions of optimal image contrast, where hypercellular and abnormal nuclear features could be identified, and would serve as diagnostic features for image classification (Fig. \ref{localizationfig}, columns 2 and 4). Additionally, sliding window method was able to identify diagnostic aggregates of abnormally large malignant glioma cells and  atypically hypercellular areas (Fig. \ref{localizationfig}, third column). 
 
 Activation of the neurons in the first convolutional layer (conv1) were found to highlight areas with increased fluorescein signal, a sign specific to brain tumor regions. Increased fluorescent signal on CLE images represent areas with blood brain barrier disruption which correspond to the tumor areas visible on a contrast enhanced MR imaging.  Interestingly, sliding window method and selected colored activation maps were not distracted or deceived by the red blood cells contamination, as they mostly highlighted tumor and brain cells rather then hypercellular areas due to bleeding. The proposed feature localization approach may be useful in the future to aid in the identification of not only the diagnostic frames, but also directing the surgeon's attention to the image parts containing major histopathological features.

\begin{table}[th!]
\centering
\caption{Interobserver study results. The model-human agreement was higher than the human-human agreement both on whole val review dataset and the gold standard subset.}
\label{interrater results}
\begin{tabular}{|c|c|c|c|}
\hline
Dataset     & \multicolumn{2}{c|}{\begin{tabular}[c]{@{}c@{}}Whole Val\\  Review\end{tabular}}                                                  & Gold-Standard                                                \\ \hline
Rater       & \begin{tabular}[c]{@{}c@{}}General \\ Agreement\end{tabular} & \begin{tabular}[c]{@{}c@{}}Cohen's \\ Kappa\end{tabular}           & \begin{tabular}[c]{@{}c@{}}General \\ Agreement\end{tabular} \\ \hline
Val-Rater 1 & 66 \%                                                        & \begin{tabular}[c]{@{}c@{}}0.32, \\ Fair\end{tabular}              & 67\%                                                         \\ \hline
Val-Rater 2 & 73 \%                                                        & \textbf{\begin{tabular}[c]{@{}c@{}}0.47, \\ Moderate\end{tabular}} & 75 \%                                                        \\ \hline
Model       & \textbf{76 \%}                                               & \textbf{\begin{tabular}[c]{@{}c@{}}0.47, \\ Moderate\end{tabular}} & \textbf{85 \%}                                               \\ \hline
\end{tabular}
\end{table}


\subsection{Inter-rater agreement}\label{Inter-rater}
Table \ref{interrater results} demonstrates the agreement between each of the val-raters and the initial review on the whole val review dataset and the gold standard subset (explained in Fig. \ref{IOfig}). The model agreement with the initial review is larger than each val-rater's agreement with the initial review. This suggests that the model has successfully learned the histological features of the CLE images that are more probable to be noticed by the neurosurgeons when the corresponding H \& E-stained histological slides were also provided for reference.

To consider images from the val review set that the majority of raters agreed on, that is one of the val-raters agreed on with the initial review, we used the gold standard subset. The gap between the model-human and human-human agreements became even more evident (19\% for val-rater 1 and 9\% for val-rater 2) with the gold standard subset (Table \ref{interrater results}, column 4).

\section{Conclusion and future work}
This paper presents a deep CNN based approach that can automatically detect the diagnostic CLE images from brain tumor surgery. We used a manually annotated in-house dataset to train and test this approach. Our results showed that both deep fine-tuning and creating an ensemble of models could enhance the performance; but only their combination could reach the maximum accuracy. The ensemble effect was stronger in DT and DFT than SFT developed models. The proposed method was also able to localize some histological features of diagnostic images. Ultimately, Table \ref{interrater results} indicates that the proposed ensemble of deeply fine-tuned models could detect the diagnostic images with a higher agreement than the trained human observers. Other confocal imaging techniques may be aided by such deep learning models. Confocal reflectance microscopy (CRM) has been studied \cite{mooney2017immediate} for rapid, fluorophore-free evaluation of pitutary adenoma biopsy specimens ex vivo. CRM allows preserving the biopsy tissue for future permanent analysis, immunohistochemical studies, and molecular studies. 
Continued use of unsupervised image segmentation methods to detect meaningful histological features from confocal brain tumor images will likely allow for more rapid and detailed diagnosis.

 \section*{Acknowledgement}
This work was supported by the Newsome Family Endowed Chair of Neurosurgery Research at the Barrow Neurological Institute held by Dr. Preul and by funds from the Barrow Neurological Foundation.


\bibliography{IEEEabrv,biblio}

\begin{thebibliography}{10}
\expandafter\ifx\csname url\endcsname\relax
  \def\url#1{\texttt{#1}}\fi
\expandafter\ifx\csname urlprefix\endcsname\relax\def\urlprefix{URL }\fi
\expandafter\ifx\csname href\endcsname\relax
  \def\href#1#2{#2} \def\path#1{#1}\fi

\bibitem{belykh2016intraoperative}
E.~Belykh, N.~L. Martirosyan, K.~Yagmurlu, E.~J. Miller, J.~M. Eschbacher,
  M.~Izadyyazdanabadi, L.~A. Bardonova, V.~A. Byvaltsev, P.~Nakaji, M.~C.
  Preul, Intraoperative fluorescence imaging for personalized brain tumor
  resection: Current state and future directions, Frontiers in Surgery 3.

\bibitem{charalampaki2015confocal}
P.~Charalampaki, M.~Javed, S.~Daali, H.-J. Heiroth, A.~Igressa, F.~Weber,
  Confocal laser endomicroscopy for real-time histomorphological diagnosis: Our
  clinical experience with 150 brain and spinal tumor cases, Neurosurgery 62
  (2015) 171--176.

\bibitem{foersch2012confocal}
S.~Foersch, A.~Heimann, A.~Ayyad, G.~A. Spoden, L.~Florin, K.~Mpoukouvalas,
  R.~Kiesslich, O.~Kempski, M.~Goetz, P.~Charalampaki, Confocal laser
  endomicroscopy for diagnosis and histomorphologic imaging of brain tumors in
  vivo, PLoS One 7~(7) (2012) e41760.

\bibitem{sanai2011intraoperative}
N.~Sanai, J.~Eschbacher, G.~Hattendorf, S.~W. Coons, M.~C. Preul, K.~A. Smith,
  P.~Nakaji, R.~F. Spetzler, Intraoperative confocal microscopy for brain
  tumors: a feasibility analysis in humans, Neurosurgery 68 (2011)
  ons282--ons290.

\bibitem{zehri2014neurosurgical}
A.~Zehri, W.~Ramey, J.~Georges, M.~Mooney, N.~Martirosyan, M.~Preul, P.~Nakaji,
  Neurosurgical confocal endomicroscopy: A review of contrast agents, confocal
  systems, and future imaging modalities, Surgical neurology international 5
  (2014) 60.

\bibitem{martirosyan2016prospective}
N.~L. Martirosyan, J.~M. Eschbacher, M.~Y.~S. Kalani, J.~D. Turner, E.~Belykh,
  R.~F. Spetzler, P.~Nakaji, M.~C. Preul, Prospective evaluation of the utility
  of intraoperative confocal laser endomicroscopy in patients with brain
  neoplasms using fluorescein sodium: experience with 74 cases, Neurosurgical
  focus 40~(3) (2016) E11.

\bibitem{reviewDLMI}
H.~Greenspan, B.~van Ginneken, R.~M. Summers, Guest editorial deep learning in
  medical imaging: Overview and future promise of an exciting new technique,
  IEEE Transactions on Medical Imaging 35~(5) (2016) 1153--1159.

\bibitem{deepmedsurvey}
G.~Litjens, T.~Kooi, B.~E. Bejnordi, A.~A.~A. Setio, F.~Ciompi, M.~Ghafoorian,
  J.~A. van~der Laak, B.~van Ginneken, C.~I. S{\'a}nchez, A survey on deep
  learning in medical image analysis, arXiv preprint arXiv:1702.05747.

\bibitem{ciresan}
D.~Ciresan, A.~Giusti, L.~M. Gambardella, J.~Schmidhuber, Deep neural networks
  segment neuronal membranes in electron microscopy images, in: Advances in
  neural information processing systems, 2012, pp. 2843--2851.

\bibitem{Zhao2016}
J.~Zhao, M.~Zhang, Z.~Zhou, J.~Chu, F.~Cao, Automatic detection and
  classification of leukocytes using convolutional neural networks, Medical \&
  biological engineering \& computing (2016) 1--15.

\bibitem{sirinukunwattana}
K.~Sirinukunwattana, S.~E.~A. Raza, Y.-W. Tsang, D.~R. Snead, I.~A. Cree, N.~M.
  Rajpoot, Locality sensitive deep learning for detection and classification of
  nuclei in routine colon cancer histology images, IEEE transactions on medical
  imaging 35~(5) (2016) 1196--1206.

\bibitem{suk2016deep}
H.-I. Suk, D.~Shen, Deep ensemble sparse regression network for alzheimer’s
  disease diagnosis, in: International Workshop on Machine Learning in Medical
  Imaging, Springer, 2016, pp. 113--121.

\bibitem{shi2017multimodal}
J.~Shi, X.~Zheng, Y.~Li, Q.~Zhang, S.~Ying, Multimodal neuroimaging feature
  learning with multimodal stacked deep polynomial networks for diagnosis of
  alzheimer's disease, IEEE journal of biomedical and health informatics.

\bibitem{salehi2017auto}
S.~S.~M. Salehi, D.~Erdogmus, A.~Gholipour, Auto-context convolutional neural
  network (auto-net) for brain extraction in magnetic resonance imaging, IEEE
  Transactions on Medical Imaging.

\bibitem{ghafoorian2017deep}
M.~Ghafoorian, N.~Karssemeijer, T.~Heskes, M.~Bergkamp, J.~Wissink, J.~Obels,
  K.~Keizer, F.-E. de~Leeuw, B.~van Ginneken, E.~Marchiori, et~al., Deep
  multi-scale location-aware 3d convolutional neural networks for automated
  detection of lacunes of presumed vascular origin, NeuroImage: Clinical 14
  (2017) 391--399.

\bibitem{zhao2016multiscale}
L.~Zhao, K.~Jia, Multiscale cnns for brain tumor segmentation and diagnosis,
  Computational and mathematical methods in medicine 2016.

\bibitem{mahapatra2016retinal}
D.~Mahapatra, P.~K. Roy, S.~Sedai, R.~Garnavi, Retinal image quality
  classification using saliency maps and cnns, in: International Workshop on
  Machine Learning in Medical Imaging, Springer, 2016, pp. 172--179.

\bibitem{purang}
A.~Abdi, C.~Luong, T.~Tsang, J.~Jue, D.~Hawley, S.~Fleming, K.~Gin, J.~Swift,
  R.~Rohling, P.~Abolmaesumi, Automatic quality assessment of echocardiograms
  using convolutional neural networks: Feasibility on the apical four-chamber
  view, IEEE Transactions on Medical Imaging.

\bibitem{gao2016d}
Y.~Gao, M.~A. Maraci, J.~A. Noble, Describing ultrasound video content using
  deep convolutional neural networks, in: Biomedical Imaging (ISBI), 2016 IEEE
  13th International Symposium on, IEEE, 2016, pp. 787--790.

\bibitem{kumar2016plane}
A.~Kumar, P.~Sridar, A.~Quinton, R.~K. Kumar, D.~Feng, R.~Nanan, J.~Kim, Plane
  identification in fetal ultrasound images using saliency maps and
  convolutional neural networks, in: Biomedical Imaging (ISBI), 2016 IEEE 13th
  International Symposium on, IEEE, 2016, pp. 791--794.

\bibitem{yosinski2014transferable}
J.~Yosinski, J.~Clune, Y.~Bengio, H.~Lipson, How transferable are features in
  deep neural networks?, in: Advances in neural information processing systems,
  2014, pp. 3320--3328.

\bibitem{Nima}
N.~Tajbakhsh, J.~Y. Shin, S.~R. Gurudu, R.~T. Hurst, C.~B. Kendall, M.~B.
  Gotway, J.~Liang, Convolutional neural networks for medical image analysis:
  full training or fine tuning?, IEEE transactions on medical imaging 35~(5)
  (2016) 1299--1312.

\bibitem{dietterich2000ensemble}
T.~G. Dietterich, et~al., Ensemble methods in machine learning, Multiple
  classifier systems 1857 (2000) 1--15.

\bibitem{zhou2002ensembling}
Z.-H. Zhou, J.~Wu, W.~Tang, Ensembling neural networks: many could be better
  than all, Artificial intelligence 137~(1-2) (2002) 239--263.

\bibitem{ciregan2012multi}
D.~Ciregan, U.~Meier, J.~Schmidhuber, Multi-column deep neural networks for
  image classification, in: Computer Vision and Pattern Recognition (CVPR),
  2012 IEEE Conference on, IEEE, 2012, pp. 3642--3649.

\bibitem{kumar2017ensemble}
A.~Kumar, J.~Kim, D.~Lyndon, M.~Fulham, D.~Feng, An ensemble of fine-tuned
  convolutional neural networks for medical image classification, IEEE journal
  of biomedical and health informatics 21~(1) (2017) 31--40.

\bibitem{christodoulidis2017multisource}
S.~Christodoulidis, M.~Anthimopoulos, L.~Ebner, A.~Christe, S.~Mougiakakou,
  Multisource transfer learning with convolutional neural networks for lung
  pattern analysis, IEEE journal of biomedical and health informatics 21~(1)
  (2017) 76--84.

\bibitem{krogh1995neural}
A.~Krogh, J.~Vedelsby, Neural network ensembles, cross validation, and active
  learning, in: Advances in neural information processing systems, 1995, pp.
  231--238.

\bibitem{izady2017improv}
M.~Izadyyazdanabadi, E.~Belykh, N.~Martirosyan, J.~Eschbacher, P.~Nakaji,
  Y.~Yang, M.~C. Preul, Improving utility of brain tumor confocal laser
  endomicroscopy: objective value assessment and diagnostic frame detection
  with convolutional neural networks, in: SPIE Medical Imaging, International
  Society for Optics and Photonics, 2017, pp. 101342J--101342J.

\bibitem{LeCun:1998:CNI:303568.303704}
Y.~LeCun, Y.~Bengio, \href{http://dl.acm.org/citation.cfm?id=303568.303704}{The
  handbook of brain theory and neural networks}, MIT Press, Cambridge, MA, USA,
  1998, Ch. Convolutional networks for images, speech, and time series, pp.
  255--258.
\newline\urlprefix\url{http://dl.acm.org/citation.cfm?id=303568.303704}

\bibitem{jia2014caffe}
Y.~Jia, E.~Shelhamer, J.~Donahue, S.~Karayev, J.~Long, R.~Girshick,
  S.~Guadarrama, T.~Darrell, Caffe: Convolutional architecture for fast feature
  embedding, arXiv preprint arXiv:1408.5093.

\bibitem{krizhevsky2012imagenet}
A.~Krizhevsky, I.~Sutskever, G.~E. Hinton, Imagenet classification with deep
  convolutional neural networks, in: Advances in neural information processing
  systems, 2012, pp. 1097--1105.

\bibitem{dropout}
N.~Srivastava, G.~Hinton, A.~Krizhevsky, I.~Sutskever, R.~Salakhutdinov,
  Dropout: A simple way to prevent neural networks from overfitting, The
  Journal of Machine Learning Research 15~(1) (2014) 1929--1958.

\bibitem{metz1978basic}
C.~E. Metz, Basic principles of roc analysis, in: Seminars in nuclear medicine,
  Vol.~8, Elsevier, 1978, pp. 283--298.

\bibitem{szegedy2015going}
C.~Szegedy, W.~Liu, Y.~Jia, P.~Sermanet, S.~Reed, D.~Anguelov, D.~Erhan,
  V.~Vanhoucke, A.~Rabinovich, Going deeper with convolutions, in: Proceedings
  of the IEEE Conference on Computer Vision and Pattern Recognition, 2015, pp.
  1--9.

\bibitem{mooney2017immediate}
M.~A. Mooney, J.~Georges, M.~I. Yazdanabadi, K.~Y. Goehring, W.~L. White, A.~S.
  Little, M.~C. Preul, S.~W. Coons, P.~Nakaji, J.~M. Eschbacher, Immediate
  ex-vivo diagnosis of pituitary adenomas using confocal reflectance
  microscopy: a proof-of-principle study, Journal of Neurosurgery.

\end{thebibliography}

\end{document}